\documentclass[twocolumn]{autart_arxiv}

\usepackage{amsmath,amssymb,amsfonts}
\usepackage{graphicx}
\usepackage{booktabs} 
\usepackage[hidelinks]{hyperref} 
\usepackage{mathtools} 
\usepackage{bm} 
\usepackage{textcomp} 
\usepackage{gensymb} 
\usepackage{svg} 

\usepackage[noadjust]{cite}

\usepackage{adjustbox}
\usepackage{array}
\newcolumntype{R}[2]{%
    >{\adjustbox{angle=#1,lap=\width-(#2)}\bgroup}%
    l%
    <{\egroup}%
}

\begin{document}

\begin{frontmatter}

\title{On the sensitivity of pose estimation neural networks: rotation parameterizations, Lipschitz constants, and provable bounds} 

\thanks{This work was supported by ONR grant N00014-17-1-2623.}
\thanks{Code is available at \url{https://github.com/uwaa-ndcl/pose_network_sensitivity}}

\author{Trevor Avant}\ead{trevoravant@gmail.com},
\author{Kristi A. Morgansen}\ead{morgansn@uw.edu}

\address{Department of Aeronautics and Astronautics, University of Washington, Seattle, WA 98195-2400, United States}  

\begin{abstract}                          
In this paper, we approach the task of determining sensitivity bounds for pose estimation neural networks.
This task is particularly challenging as it requires characterizing the sensitivity of 3D rotations.
We develop a sensitivity measure that describes the maximum rotational change in a network's output with respect to a Euclidean change in its input.
We show that this measure is a type of Lipschitz constant, and that it is bounded by the product of a network's Euclidean Lipschitz constant and an intrinsic property of a rotation parameterization which we call the ``distance ratio constant''.
We derive the distance ratio constant for several rotation parameterizations, and then discuss why the structure of most of these parameterizations makes it difficult to construct a pose estimation network with provable sensitivity bounds.
However, we show that sensitivity bounds can be computed for networks which parameterize rotation using unconstrained exponential coordinates.
We then construct and train such a network and compute sensitivity bounds for it.
\end{abstract}

\end{frontmatter}

\section{Introduction}

In the past few years, deep neural networks have achieved remarkable success at the task of object pose estimation \cite{Rad,Tremblay,Xiang,Tekin}.
However, a major downside of deep learning-based methods is that deep neural networks are not well-understood mathematically and lack provable performance guarantees.
This lack of theoretical understanding is problematic as pose estimation networks are often intended to be applied to real-world applications which require safe and reliable operation.
Therefore, it is necessary to develop better tools to analyze pose estimation networks in order to use them in the real world.


One significant disadvantage of deep neural networks is that they are often highly sensitive, and small changes in a network's input can result in huge changes in its output \cite{Szegedy}.
One of the primary tools for analyzing neural network sensitivity is the Lipschitz constant, which measures the maximum distance between two outputs of a function with respect to the distance between the two corresponding inputs, taken over all admissible pairs of inputs.
Although exact computation of a network's Lipschitz constant is too formidable for all but very small networks \cite{Jordan}, an upper bound can be computed instead.
Other work has also considered estimating Lipschitz constants \cite{Scaman,Latorre,Fazlyab,Zou},
as well as specifically analyzing the sensitivity of classification networks (often with regard to adversarial examples) \cite{Tjeng,Peck,Tsuzuku,Weng}.

While sensitivity analysis of general neural networks and of classification neural networks are active areas of research, to the best of our knowledge, the sensitivity of pose estimation neural networks has not yet been studied.
This task is particularly challenging as sensitivity must be considered using a measure of rotational distance rather than a norm-based metric such as the Euclidean distance.
As such, we measure network sensitivity as the maximum rotational distance between two outputs of a network with respect to the Euclidean distance between the two corresponding inputs.
We show that this measure is a type of Lipschitz constant, and that it is bounded by the product of two values: the network's Euclidean Lipschitz constant, and a measure we call the ``distance ratio constant'' which is the maximum ratio of the rotational and Euclidean distances between all pairs of rotation parameters.

Analyzing the sensitivity of a pose estimation network will depend on which specific rotation parameterization is used.
A variety of rotation parameterizations have been used in the literature, with some examples shown in Figure \ref{fig:pose_estimation_networks_table}.
In this paper we consider five types of rotation parameterizations: exponential coordinates, unconstrained exponential coordinates, quaternions, unconstrained quaternions, and 2D projection parameterizations.
We show how the structure of most of these parameterizations raise difficulties in constructing a pose estimation network with provable rotational sensitivity bounds.
However, we demonstrate that such bounds can be determined for networks which parameterize rotation using unconstrained exponential coordinates.
We then construct such a network and calculate rotational sensitivity bounds for it.


\begin{figure}[ht]
\setlength{\tabcolsep}{4.8pt}
\centering
\begin{tabular}{ l l l l l }
\toprule
\textbf{Network} & \textbf{Authors} & \textbf{Year} & \textbf{Param.} & \textbf{Ref.} \\
\midrule
PoseNet & Kendall et al. & 2015 & quat. & \cite{Kendall} \\
BB8 & Rad et al. & 2017 & 2D proj & \cite{Rad} \\
DOPE & Tremblay et al. & 2018 & 2D proj & \cite{Tremblay} \\
PoseCNN & Xiang et al. & 2018 & quat. & \cite{Xiang} \\
- & Tekin et al. & 2018 & 2D proj & \cite{Tekin} \\
\bottomrule
\end{tabular}
\caption{
Several of the earliest pose estimation neural networks, along with the rotation parameterization used in each method.
In this table ``quat.'' denotes ``quaternions'', and the term ``2D proj'' denotes 2D projection parameterizations, in which pose is represented by projecting 3D points on an object into the 2D image plane.
}
\label{fig:pose_estimation_networks_table}
\end{figure}



The remainder of this paper is organized as follows.
In Section \ref{sec:pose}, we discuss mathematical properties of 3D rotations and pose.
In Section \ref{sec:lipschitz_constants_and_pose_estimation}, we discuss Lipschitz constants, construct a Lipschitz constant that measures the sensitivity of a pose estimation neural network and derive a bound on it.
We derive the distance ratio constant for exponential coordinates and quaternions, respectively, in Sections \ref{sec:exponenential_coordinates} and \ref{sec:quaternions}.
In Section \ref{sec:summary_and_simulation}, we summarize our findings from the previous sections and discuss their implications in creating a pose estimation neural network with provable bounds.
We then go on to construct a pose estimation neural network which outputs rotation in the form of unconstrained exponential coordinates, and calculate sensitivity bounds for the network.
Finally, we discuss areas of future work and provide concluding remarks in Section \ref{sec:conclusion}.


















\section{Pose and 3D rotations} \label{sec:pose}

In this section, we will provide background information on pose and 3D rotations of a rigid objects, and we will define a distance measure between 3D rotations.

\subsection{Pose and 3D rotations}

The pose of a rigid object is defined as its position and orientation.
The set of all 3D positions and orientations is topologically characterized by $\textrm{SE(3)}$, the ``special Euclidean group'' in three dimensional Euclidean space.
Pose is often parameterized by treating position and rotation separately, in which case position can be described by a 3D vector, and rotation can be described by a rotation parameterization such as exponential coordinates or quaternions.
In other representations of pose, such as 2D projected points, position and rotation are coupled.

The set of all rotations about the origin in three-dimensional Euclidean space is termed the special orthogonal group in three dimensions, $\textrm{SO(3)}$. 
Each element of $\textrm{SO(3)}$ can be described by an orthogonal $3{\times}3$ matrix with determinant equal to one:
\begin{equation}
\textrm{SO(3)} = \{ \mathbf{R} \in \mathbb{R}^{3 \times 3} ~~ | ~~ \mathbf{R}^T \mathbf{R} = \mathbf{I}, ~~ \textrm{det}(\mathbf{R}) = 1 \}
\end{equation}
where $\textrm{det}(\cdot)$ denotes the determinant operation, and $\mathbf{I} \in \mathbb{R}^{3 \times 3}$ denotes the identity matrix.

There are many different parameterizations of the group of 3D rotations, including quaternions, exponential coordinates, and Euler angles.
In this paper, we will use the symbol $\mathcal{P} \subseteq \mathbb{R}^m$ to denote a set of rotation parameters, and the symbol $\mathbf{p} \in \mathcal{P}$ to denote a particular parameter in that set.
The symbols $\mathcal{P}$ and $\mathbf{p}$ are generic in that they do not correspond to any specific parameterization, and they will be replaced with specific symbols when considering a specific rotation parameterization.
For example, when considering quaternions, $\mathcal{P}$ will be replaced with $\mathcal{S}^3 \subset \mathbb{R}^4$ (see \eqref{eq:S3}), and $\mathbf{p}$ will be replaced with $\mathbf{q} \in \mathbb{R}^4$.

\subsection{Rotational distance}

We now consider the task of quantifying the ``distance'' between two 3D rotations.
As the set of 3D rotations is structured as a Riemannian manifold, Euclidean distance cannot be used to properly measure distance for rotations.
A proper measure of rotational distance is the angle of direct rotation between two 3D rotations.
This measure provides a consistent measure of rotational distance, can be applied to any rotation parameterization, and is intuitive as it corresponds to an angle of rotation.
\begin{defn} \label{defn:rotational_distance}
The \textbf{rotational distance} between two 3D rotations is the angle of direct rotation between the two rotations using an axis-angle representation.
Letting $\mathcal{P}$ denote a set of rotation parameters, we denote the rotational distance function as $\mathrm{dist}: \mathcal{P} \times \mathcal{P} \rightarrow [0,\pi]$.
\end{defn}


The function $\mathrm{dist}(\cdot,\cdot)$ is overloaded notation as its explicit expression will depend on which rotation parameterization is being considering.
For example, the rotational distance between two rotation matrices $\mathbf{R}_1, \mathbf{R}_2 \in \mathbb{R}^{3 \times 3}$ is given by the following equation:
\begin{align}
\mathrm{dist}(\mathbf{R}_1, \mathbf{R}_2)
&= \cos^{-1} \left( \tfrac{1}{2} (\textrm{tr}(\mathbf{R}_1 \mathbf{R}_2^T) - 1) \right)
\end{align}
where $\textrm{tr}(\cdot)$ denotes the trace operation.
Later in the paper, we consider the distance between two pairs of exponential coordinates, $\mathrm{dist}(\mathbf{s}_1,\mathbf{s}_2)$, in \eqref{eq:dist_axang}, and the distance between two quaternions, $\mathrm{dist}(\mathbf{q}_1,\mathbf{q}_2)$, in \eqref{eq:dist_quat}.


\section{Lipschitz constants and measures of sensitivity for pose estimation neural networks} \label{sec:lipschitz_constants_and_pose_estimation}

In this section, we will consider measuring the sensitivity of a pose estimation network, which can be thought of as a mathematical function which inputs a camera image (or similar structure) and outputs a rotation parameterization.
Our analysis requires several tools which we will now define: metrics, pseudometrics, and Lipschitz constants.


\subsection{Lipschitz constants} \label{sec:lipschitz_constants}

We start by defining metrics and pseudometrics \cite[Ch.~4]{Kelley}, which will be used to define Lipschitz constants.
\begin{defn} \label{defn:metric}
A function $d: \mathcal{X} \times \mathcal{X} \to [0, \infty)$ is a \textbf{metric} on set $\mathcal{X}$ if it satisfies properties (a), (b), (c), and (d) below, for all $x,y,z \in \mathcal{X}$.
Additionally, a function $d: \mathcal{X} \times \mathcal{X} \to [0, \infty)$ is a \textbf{pseudometric} on set $\mathcal{X}$ if it only satisfies properties (a), (b), and (c), for all $x,y,z \in \mathcal{X}$.
\begin{enumerate}
\item[(a)] $d(x,y) = d(y,x)$
\item[(b)] $d(x,y) \leq  d(x,z) + d(z,y)$
\item[(c)] $d(x,y) = 0 ~~~ \Leftarrow ~~~ x=y$
\item[(d)] $d(x,y) = 0 ~~~ \Rightarrow ~~~ x=y$
\end{enumerate}
\end{defn}

Lipschitz constants are a tool to quantify the sensitivity of mathematical functions, and are one of the main tools used to analyze the sensitivity of neural networks.
Given a function, $\mathbf{f}$, the Lipschitz constant describes the maximum amount the output of the function can change with respect to changes in the input \cite[Ch.~9]{Searcoid}.
Lipschitz constants are useful due to their applicability to a wide array of functions, including vector-valued, nonlinear, and non-differentiable functions.
\begin{defn} \label{defn:lipschitz_constant}
Consider a function $\mathbf{f}: \mathcal{X} \rightarrow \mathcal{Y}$, and metrics $d_{\mathcal{X}}: \mathcal{X} \times \mathcal{X} \rightarrow [0, \infty)$ and $d_{\mathcal{Y}}: \mathcal{Y} \times \mathcal{Y} \rightarrow [0, \infty)$.
The \textbf{Lipschitz constant}, $L \in \mathbb{R}$, of $\mathbf{f}$ with respect to $d_{\mathcal{X}}$ and $d_{\mathcal{Y}}$ is the minimal $L \geq 0$ such that
\begin{equation}
d_{\mathcal{Y}}(\mathbf{f}(\mathbf{x}_1), \mathbf{f}(\mathbf{x}_2)) \leq L d_{\mathcal{X}}(\mathbf{x}_1, \mathbf{x}_2), ~~~~ \forall \mathbf{x}_1,\mathbf{x}_2 \in \mathcal{X} .
\label{eq:global_lipschitz_constant_original}
\end{equation}
\end{defn}

Note that some authors define any $L$ that satisfies the inequality above as ``a Lipschitz constant'', but we will define ``the Lipschitz constant'' as the minimal $L$ for which this inequality holds, and we refer to any larger value as an upper bound. We can solve for $L$ in \eqref{eq:global_lipschitz_constant_original} as
\begin{equation}
L = \sup_{\substack{\mathbf{x}_1, \mathbf{x}_2 \in \mathcal{X} \\ \mathbf{x}_1 \neq \mathbf{x}_2}} \frac{d_{\mathcal{Y}}(\mathbf{f}(\mathbf{x}_1), \mathbf{f}(\mathbf{x}_2))}{d_{\mathcal{X}}(\mathbf{x}_1, \mathbf{x}_2)} .
\label{eq:global_lipschitz_constant}
\end{equation}
Note that excluding points such that $\mathbf{x}_1{=}\mathbf{x}_2$ does not affect the supremization above since these points satisfy \eqref{eq:global_lipschitz_constant_original} for any $L$.

\subsection{Euclidean Lipschitz constants} \label{sec:euclidean_lipschitz_constants}

In this paper, we are interested in the case in which the function $\mathbf{f}$ is a neural network that maps an image or other type of input to a rotation parameter.
We can write this function as $\mathbf{f}: \mathbb{R}^n \to \mathcal{P}$.
Lipschitz constants are typically applied to neural networks using the metrics induced by the 1-, 2-, and $\infty$-norms for $d_{\mathcal{X}}$ and $d_{\mathcal{Y}}$.
We now present the Lipschitz constant in which both metrics are defined as the Euclidean distance.
\begin{defn}
Consider a function $\mathbf{f}: \mathbb{R}^n \rightarrow \mathcal{P}$ which outputs a rotation parameter.
The \textbf{Euclidean Lipschitz constant}, $L_e$, of $\mathbf{f}$ is the Lipschitz constant of the function using the Euclidean distance as the metric on both the input and output sets:
\begin{equation} \label{eq:euclidean_lipschitz_constant}
L_e \coloneqq \sup_{\mathbf{x}_1 \neq \mathbf{x}_2} \frac{\lVert \mathbf{f}(\mathbf{x}_2) - \mathbf{f}(\mathbf{x}_1) \rVert}{\lVert \mathbf{x}_2 - \mathbf{x}_1 \rVert} .
\end{equation}
\end{defn}

\subsection{Rotational Lipschitz constants} \label{sec:rotational_lipschitz_constants}

Considering a neural network, $\mathbf{f}$, that maps an image to a rotation parameter, the Euclidean Lipschitz constant does not provide a meaningful measure of sensitivity due to the fact that the Euclidean distance between two rotation parameters is not a meaningful measure of rotational distance.
So, to properly measure the sensitivity of these functions, we will create a Lipschitz constant that measures the distance between inputs using the Euclidean distance, and the distance between outputs using the rotational distance from Definition \ref{defn:rotational_distance}.
We present this measure in the following definition.

\begin{defn} \label{defn:rotational_lipschitz_constant}
Consider a function $\mathbf{f}: \mathbb{R}^n \rightarrow \mathcal{P}$ that outputs a rotation parameter.
The \textbf{rotational Lipschitz constant}, $L_r$, of $\mathbf{f}$ is the Lipschitz constant of the function using the Euclidean distance as the metric on the input set, and the rotational distance as the metric on the output set:
\begin{equation} \label{eq:rotational_lipschitz_constant}
L_r \coloneqq
\sup_{\mathbf{x}_1 \neq \mathbf{x}_2} \frac{\mathrm{dist}(\mathbf{f}(\mathbf{x}_1), \mathbf{f}(\mathbf{x}_2))}{\lVert \mathbf{x}_2 - \mathbf{x}_1 \rVert} .
\end{equation}
\end{defn}

We now note that the rotational distance function $\mathrm{dist}(\cdot,\cdot)$ from Definition \ref{defn:rotational_distance} is a pseudometric and not a metric, which is due to the fact for some rotation parameterizations, $\mathrm{dist}(\mathbf{p}_1,\mathbf{p}_2)$ may equal zero for some parameters $\mathbf{p}_1{\neq}\mathbf{p}_2$.
As a result, the rotational Lipschitz constant $L_r$ is not a true Lipschitz constant as described in Definition \ref{defn:lipschitz_constant}.
However, this discrepancy is minor and does not have any negative effects on using the rotational Lipschitz constant to measure the sensitivity of a neural network.

As with Euclidean Lipschitz constants, it is too difficult to exactly calculate the rotational Lipschitz constant of complex functions such as neural networks, so instead we will derive an upper bound.
This bound will require defining the following measure which relates the Euclidean and rotational distances between pairs of rotation parameters.

\begin{defn} \label{defn:distance_ratio_constant}
Consider a rotation parameterization with set of parameters, $\mathcal{P}$.
The \textbf{distance ratio constant}, $\mu$, of the parameterization is the maximum ratio of the rotational distance to the Euclidean distance over all pairs of parameters:
\begin{equation}
\mu \coloneqq \sup_{\substack{\mathbf{p}_1, \mathbf{p}_2 \in \mathcal{P} \\ \mathbf{p}_1 \neq \mathbf{p}_2}} \frac{\mathrm{dist}(\mathbf{p}_1, \mathbf{p}_2)}{\lVert \mathbf{p}_2 - \mathbf{p}_1 \rVert} .
\label{eq:rotational_sensitivity_constant}
\end{equation}
\end{defn}

Note this definition is equivalent to applying the rotational Lipschitz constant from Definition \ref{defn:rotational_lipschitz_constant} to the identity map $\mathbf{f}(\mathbf{p}) = \mathbf{p}$.

\subsection{Bounds on the rotational Lipschitz constant}

Using $L_e$ and $\mu$, we can now derive a bound on a rotational Lipschitz constant $L_r$.

\begin{thm} \label{thm:network_wide_bound}
Consider a function $\mathbf{f}: \mathbb{R}^n \rightarrow \mathcal{P}$ that outputs a rotation parameter.
Let $L_r$ denote the rotational Lipschitz constant of $\mathbf{f}$, let $L_e$ denote the Euclidean Lipschitz constant of $\mathbf{f}$, and let $\mu$ denote the distance ratio constant of the rotation parameterization.
The following equation holds:
\begin{equation} \label{eq:network_wide_bound}
L_r
\leq
\mu
L_e .
\end{equation}
\end{thm}
\begin{pf}
Using the definition of the rotational Lipschitz constant in \eqref{eq:rotational_lipschitz_constant}, noting that $\mathbf{f}(\mathbf{x}_1)=\mathbf{p}_1$ and $\mathbf{f}(\mathbf{x}_2)=\mathbf{p}_2$, and applying the definitions of $L_e$ and $\mu$ from \eqref{eq:euclidean_lipschitz_constant} and \eqref{eq:rotational_sensitivity_constant}, we have
\begin{align}
L_r &= \sup_{\mathbf{x}_1 \neq \mathbf{x}_2} \frac{\mathrm{dist}(\mathbf{p}_1, \mathbf{p}_2)}{\lVert \mathbf{x}_2 - \mathbf{x}_1 \rVert} \\
&\hspace{0mm} \leq
\sup_{\mathbf{x}_1 \neq \mathbf{x}_2} \frac{\lVert \mathbf{p}_2 - \mathbf{p}_1 \rVert}{\lVert \mathbf{x}_2 - \mathbf{x}_1 \rVert}
\sup_{\substack{\mathbf{p}_1, \mathbf{p}_2 \in \mathcal{P} \\ \mathbf{p}_1 \neq \mathbf{p}_2}} \frac{\mathrm{dist}(\mathbf{p}_1, \mathbf{p}_2)}{\lVert \mathbf{p}_2 - \mathbf{p}_1 \rVert} \\
&\hspace{0mm} = L_e
\mu .
\end{align}
\\[-6mm]\hspace*{\fill}$\square$
\end{pf}


Theorem \ref{thm:network_wide_bound} can be used in practice to determine rotational bounds on the output of a function based on Euclidean bounds on the input.
This result is summarized in the following corollary.
\begin{cor} \label{cor:distance_bound}
Consider a function $\mathbf{f}: \mathbb{R}^n \rightarrow \mathcal{P}$ that outputs a rotation parameter.
Let $L_e$ denote the Euclidean Lipschitz constant of the function, and let $\mu$ denote the distance ratio constant of the rotation parameterization.
Let $\mathbf{x}_1, \mathbf{x}_2 \in \mathbb{R}^n$ denote two inputs to the function, and let $\mathbf{p}_1, \mathbf{p}_2 \in \mathcal{P}$ denote the corresponding outputs.
If $\lVert \mathbf{x}_2 - \mathbf{x}_1 \rVert \leq \epsilon$ for some $\epsilon \geq 0$, then the rotational distance between the outputs
is bounded as follows:
\begin{align}
\mathrm{dist}(\mathbf{p}_1, \mathbf{p}_2)
&\leq
\epsilon
\mu
L_e .
\label{eq:distance_bound}
\end{align}
\end{cor}
\begin{pf}
Note that \eqref{eq:network_wide_bound} implies
\begin{align}
\begin{aligned}
\frac{\mathrm{dist}(\mathbf{p}_1, \mathbf{p}_2)}{\lVert \mathbf{x}_2 - \mathbf{x}_1 \rVert}
&\leq
L_e
\mu,
& &\forall \mathbf{x}_1 \neq \mathbf{x}_2 \\
\mathrm{dist}(\mathbf{p}_1, \mathbf{p}_2)
&\leq
\lVert \mathbf{x}_2 - \mathbf{x}_1 \rVert
L_e
\mu,
& &\forall \mathbf{x}_1 \neq \mathbf{x}_2 .
\end{aligned}
\end{align}
Combining $\lVert \mathbf{x}_2 - \mathbf{x}_1 \rVert \leq \epsilon$ with the equation above yields \eqref{eq:distance_bound}. 
\hspace*{\fill}$\square$
\end{pf}

Note that this bound will only be useful if the right hand side of \eqref{eq:distance_bound} is less than the maximum possible rotational distance of $\pi$.

\subsection{Application to pose estimation neural networks}

The function, $\mathbf{f}$, that we have considered in this section will correspond to a feedforward pose estimation neural network.
In order to apply the results from this section to such a network, there are several considerations we must make.

First, note that for networks which take RGB images as their inputs, the input, $\mathbf{x}$, will be a three-dimensional array.
It is mathematically equivalent to consider such inputs to be vectors in $\mathbb{R}^n$.
Also note that we have considered a function, $\mathbf{f}$, which outputs rotation parameters, but a pose estimation neural network will output pose, which consists of both position and rotation.
However, pose is often represented as a concatenation of position and rotation, so in these cases we can split a network into position and rotation sub-networks, and consider each sub-network separately.
Finally, as we noted before, the exact Euclidean Lipschitz constant, $L_e$, is intractable to compute for all but very small toy networks, but an upper bound can be computed instead by taking the product of the individual bounds of the layers of the network.

In summary, we would like to compute a bound on the rotational Lipschitz constant from Theorem \ref{thm:network_wide_bound} for a pose estimation neural network.
This bound is a function of two values: the Euclidean Lipschitz constant, $L_e$, and the distance ratio constant, $\mu$.
While we know how to compute an upper bound on $L_e$, we are not aware of any work in which $\mu$ is derived for any rotation parameterization.
Accordingly, we will devote the next two sections of the paper to the derivation of $\mu$ for two types of rotation parameterizations: exponential coordinates and quaternions.

Note that a common pose parameterization used in pose estimation networks is 2D projected points, in which pose is parameterized by a set of 2D points which represent 3D points on the object which have been projected into the image plane \cite{Rad,Tremblay,Tekin}.
Unfortunately, for these types of parameterizations, the mapping between parameters and pose relies on Perspective-n-Point (PnP) algorithms, which often are solved via nonlinear optimization and also often rely on additional algorithms such as Random Sample Consensus (RANSAC).
As a result, deriving closed-form sensitivity bounds for these parameterizations is intractable.

\section{Exponential coordinates} \label{sec:exponenential_coordinates}

In this section, we will derive the distance ratio constant from Definition \ref{defn:distance_ratio_constant} for exponential coordinates and unconstrained exponential coordinates.

\subsection{Rotation parameterization}

The first rotation parameterization we will consider is exponential coordinates.
Although not commonly used in pose estimation neural networks, exponential coordinates are a fundamental and intuitive description of rotation used frequently in robotics applications.
Additionally, as we will show later on, exponential coordinates have several properties which make them advantageous for developing pose estimation networks with provable sensitivity bounds.

Any 3D rotation can be described by a rotation of angle $\theta \in \mathbb{R}$ about a unit-length axis $\mathbf{e} \in \mathbb{R}^3$. Exponential coordinates, which we will denote as $\mathbf{s} \in \mathbb{R}^3$, are simply the axis multiplied by the angle: $\mathbf{s} = \theta \mathbf{e} \in \mathbb{R}^3$ \cite[Ch.~3]{Lynch}. The term ``exponential'' comes from the fact that $\textrm{SO(3)}$ is a Lie group, and exponential coordinates are a description of the Lie algebra of the Lie group, which can be mapped back to the group using the exponential map.

In general, exponential coordinates can be defined using any angle $\theta \in [0,\infty)$.
We refer to the set of exponential coordinates defined with $\theta \in [0,\infty)$ as ``unconstrained'' exponential coordinates.
However, exponential coordinates are typically defined with $\theta \in [0,\pi]$, which is the minimum set of angles that can describe all 3D rotations.
This set of exponential coordinates corresponds to the three-dimensional ball of radius $\pi$,
which we denote as $\mathcal{B}_{\pi}$:
\begin{align} \label{eq:Bpi}
\mathcal{B}_{\pi} = \{ \mathbf{x} \in \mathbb{R}^3 ~~ | ~~ \lVert \mathbf{x} \rVert \leq \pi \} .
\end{align}

There is a 1-to-1 mapping between points inside $\mathcal{B}_{\pi}$ and 3D rotations, and a 2-to-1 mapping between points on the surface $\mathcal{B}_{\pi}$ to 3D rotations. More specifically, antipodal points on the surface of the ball correspond to the same rotation (i.e., $\pi \mathbf{e}$ and $-\pi \mathbf{e}$ correspond to the same rotation) \cite[Ch.~3]{Lynch}.
Note that while we could reduce the set $\mathcal{B}_{\pi}$ so that it is a 1-to-1 mapping, it would not change the results in this section.

We now consider how exponential coordinates can be composed.
Let $\mathbf{s}_1 = \theta_1 \mathbf{e}_1$ and $\mathbf{s}_2 = \theta_2 \mathbf{e}_2$ denote two sets of exponential coordinates, and let $\mathbf{s}_3 = \theta_3 \mathbf{e}_3$ denote the coordinates of the composite rotation of $\mathbf{s}_1$ and $\mathbf{s}_2$.
The angle $\theta_3$ can be determined using the following formula \cite{Altmann}:
\begin{align}
\cos \tfrac{\theta_3}{2} = \cos \tfrac{\theta_1}{2} \cos \tfrac{\theta_2}{2} - \mathbf{e}_1 \cdot \mathbf{e}_2 \sin \tfrac{\theta_1}{2} \sin \tfrac{\theta_2}{2} .
\label{eq:axang_composite_rotation}
\end{align}

\subsection{Rotational distance}

Using \eqref{eq:axang_composite_rotation}, we can derive the rotational distance between two pairs of exponential coordinates.
\begin{prop} \label{prp:dist_exp_coords}
The rotational distance between two pairs of exponential coordinates, $\mathbf{s}_1 = \theta_1 \mathbf{e}_1$ and $\mathbf{s}_2 = \theta_2 \mathbf{e}_2$, is given by the following equation:
\begin{align}
\begin{aligned}
&\mathrm{dist}(\mathbf{s}_1, \mathbf{s}_2) \\
& \hspace{3mm} = 2 \cos^{-1} ( \lvert\cos \tfrac{\theta_1}{2} \cos \tfrac{\theta_2}{2} + \mathbf{e}_1 \cdot \mathbf{e}_2 \sin \tfrac{\theta_1}{2} \sin \tfrac{\theta_2}{2}\rvert ) .
\label{eq:dist_axang}
\end{aligned}
\end{align}
\end{prop}
\begin{pf}
The rotational distance can be determined by finding the angle corresponding to the composite axis-angle rotation between the first rotation $\mathbf{s}_1$ and the \textit{inverse} of the second rotation. The inverse of the second rotation is found by simply negating the angle of rotation $\theta_2$, and is therefore described by the exponential coordinates $-\mathbf{s}_2 = - \theta_2 \mathbf{e}_2$. To find the angle of the composite rotation, we can use \eqref{eq:axang_composite_rotation}:
\begin{align}
\cos \tfrac{\theta_3}{2} &= \cos \tfrac{\theta_1}{2} \cos \tfrac{-\theta_2}{2} - \mathbf{e}_1 \cdot \mathbf{e}_2 \sin \tfrac{\theta_1}{2} \sin \tfrac{-\theta_2}{2} \\
&= \cos \tfrac{\theta_1}{2} \cos \tfrac{\theta_2}{2} + \mathbf{e}_1 \cdot \mathbf{e}_2 \sin \tfrac{\theta_1}{2} \sin \tfrac{\theta_2}{2}
\label{eq:cos_theta3_over_2}
\end{align}
where we have used the trigonometric identities $\sin(-x) = - \sin(x)$ and $\cos(-x) = \cos(x)$. Next, for convenience, we define
\begin{equation}
a \coloneqq \cos \tfrac{\theta_1}{2} \cos \tfrac{\theta_2}{2} + \mathbf{e}_1 \cdot \mathbf{e}_2 \sin \tfrac{\theta_1}{2} \sin \tfrac{\theta_2}{2}
\label{eq:a}
\end{equation}
and note that $a \in [-1,1]$.
Solving for $\theta_3$ in \eqref{eq:cos_theta3_over_2} and using \eqref{eq:a}, we have
\begin{align}
\theta_3 = 2 \cos^{-1} ( a ) .
\label{eq:theta3}
\end{align}
Arccosine will map to the interval $[0,\pi]$, so $\theta_3$ in \eqref{eq:theta3} will be mapped to the interval $[0,2\pi]$. However, we would like to express $\theta_3$ as the rotational distance from zero, which is on the interval $[0,\pi]$. So, we can write
\begin{align}
\mathrm{dist}(\mathbf{s}_1, \mathbf{s}_2) =
\begin{cases}
\theta_3, & \theta_3 \in [0, \pi] \\
2 \pi - \theta_3, & \theta_3 \in [\pi, 2\pi] .
\end{cases}
\label{eq:axang_dist_cases}
\end{align}
Because $\theta_3$ will be greater than $\pi$ when the argument to the arccosine (i.e., $a$) is negative, using \eqref{eq:theta3} we can rewrite \eqref{eq:axang_dist_cases} as
\begin{align}
\mathrm{dist}(\mathbf{s}_1, \mathbf{s}_2) &=
\begin{cases}
2 \cos^{-1} (a), & a \in [0,1] \\
2 \pi - 2 \cos^{-1} (a), & a \in [-1,0] .
\end{cases}
\label{eq:axang_dist_cases2}
\end{align}
Now note that we can write $\cos^{-1}(\lvert x \rvert)$ in the following way, based on the sign of the argument:
\begin{equation}
\cos^{-1}(\lvert x \rvert) =
\begin{cases}
\pi - \cos^{-1}(x), &x \in [-1, 0] \\
\cos^{-1}(x), &x \in [0, 1] .
\end{cases}
\label{eq:arccos_abs_cases}
\end{equation}
Using \eqref{eq:arccos_abs_cases} we can write \eqref{eq:axang_dist_cases2} as \eqref{eq:dist_axang}, which completes the proof.
\hspace*{\fill}$\square$
\end{pf}

\subsection{Planar analysis} \label{sec:exp_coords_2d}

\begin{figure}[ht]
\centering
\includegraphics[width=.40\textwidth]{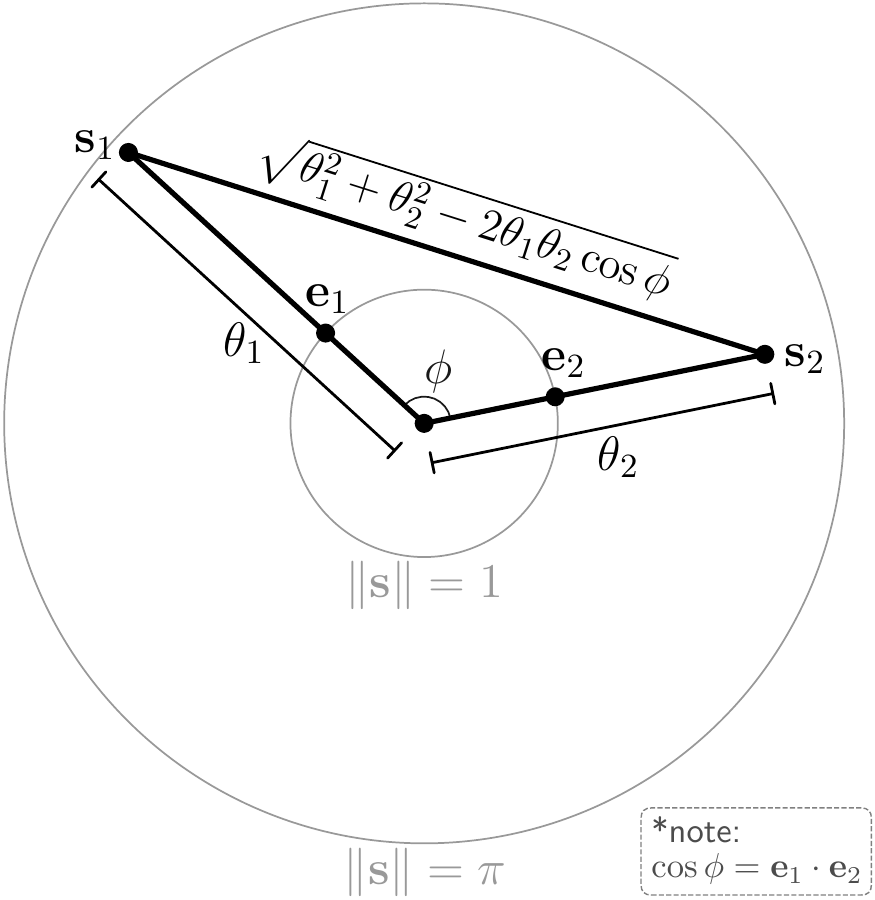}
\caption{Exponential coordinates $\mathbf{s}_1$ and $\mathbf{s}_2$ shown in the plane they define.
In the plane, two exponential coordinate vectors can be analyzed in terms of the angle $\phi$ between the vectors, and the magnitudes $\theta_1$ and $\theta_2$ of the vectors.
The Euclidean distance between the exponential coordinates, $\lVert \mathbf{s}_2 - \mathbf{s}_1 \rVert = (\theta_1^2 + \theta_2^2 - 2 \theta_1 \theta_2 \cos \phi)^{1/2}$, is computed using the law of cosines.}
\label{fig:exp_coords_2d}
\end{figure}

To determine the distance ratio constant of exponential coordinates from Definition \ref{defn:distance_ratio_constant}, we would need to optimize over all possible pairs of exponential coordinates, where each pair consists of the six degrees of freedom in $\mathbf{s}_1$ and $\mathbf{s}_2$, and the two constraints $\lVert \mathbf{s}_1 \rVert \leq \pi$ and $\lVert \mathbf{s}_2 \rVert \leq \pi$.
The number of dimensions and constraints in this analysis is challenging, but we can analyze this problem in the plane defined by $\mathbf{s}_1$ and $\mathbf{s}_2$, which reduces the number of variables and simplifies the analysis.

\begin{lem} \label{lem:exp_coords_2d}
Consider the exponential coordinates $\mathbf{s}_1 = \theta_1 \mathbf{e}_1$ and $\mathbf{s}_2 = \theta_2 \mathbf{e}_2$ where $\theta_1, \theta_2 \in [0,\infty)$. Define the variable $t$:
\begin{align}
t \coloneqq \tfrac{1}{2} (1 - \mathbf{e}_1 \cdot \mathbf{e}_2),  ~~~~~~ t \in [0, 1] .
\label{eq:e1_e2_to_t}
\end{align}
The Euclidean and rotational distances between $\mathbf{s}_1$ and $\mathbf{s}_2$ are:
\begin{align}
&\lVert \mathbf{s}_2 - \mathbf{s}_1 \rVert = \sqrt{(\theta_1 - \theta_2)^2 + 4 \theta_1 \theta_2 t} \label{eq:euclidean_distance_exp_coords_2d} \\
\begin{split}\label{eq:rotational_distance_exp_coords_2d}
&\mathrm{dist}(\mathbf{s}_1, \mathbf{s}_2) \\
&\hspace{3.0mm} = 2 \cos^{-1} ( | t \cos (\tfrac{\theta_1}{2}{+}\tfrac{\theta_2}{2}) + (1{-}t) \cos (\tfrac{\theta_1}{2}{-}\tfrac{\theta_2}{2}) | ) .
\end{split}
\end{align}
\end{lem}
\begin{pf}
We note that any two exponential coordinates $\mathbf{s}_1$ and $\mathbf{s}_2$ lie in a plane, and hence we can analyze their relationship in that plane. A diagram is shown in Fig. \ref{fig:exp_coords_2d}.

First, note the following equation which relates the dot product and angle between two vectors:
\begin{align}
\cos \phi = \mathbf{e}_1 \cdot \mathbf{e}_2 .
\label{eq:cos_phi_e1_e2}
\end{align}
To show \eqref{eq:euclidean_distance_exp_coords_2d}, we can use the law of cosines to express the Euclidean distance between two points (see Fig. \ref{fig:exp_coords_2d}):
\begin{equation}
\lVert \mathbf{s}_2 - \mathbf{s}_1 \rVert = \sqrt{\theta_1^2 + \theta_2^2 - 2 \theta_1 \theta_2 \cos \phi} .
\end{equation}
Using \eqref{eq:cos_phi_e1_e2} and \eqref{eq:e1_e2_to_t}, the equation above can be manipulated to yield \eqref{eq:euclidean_distance_exp_coords_2d}.

To show \eqref{eq:rotational_distance_exp_coords_2d}, we can substitute \eqref{eq:cos_phi_e1_e2} into \eqref{eq:dist_axang}, apply the product-to-sum formula, and then manipulate the resulting equation algebraically which yields \eqref{eq:rotational_distance_exp_coords_2d}.
\hspace*{\fill}$\square$
\end{pf}

\subsection{Distance ratio constants}

We now approach the problem of analytically determining the distance ratio constant for exponential coordinates.
We will start by deriving the distance ratio constant for exponential coordinates with angles on the interval $[0,2\pi]$.
We will then use this result to determine the distance ratio constant for unconstrained exponential coordinates and exponential coordinates in Theorems \ref{thm:drc_uncon_exp_coords} and \ref{thm:drc_exp_coords}.
\begin{lem} \label{lem:drc_exp_coords_0_2pi}
The distance ratio constant, $\mu$, for exponential coordinates $\mathbf{s}_1 = \theta_1 \mathbf{e}_1$ and $\mathbf{s}_2 = \theta_2 \mathbf{e}_2$ for which $\theta_1, \theta_2 \in [0,2\pi]$ is:
\begin{align}
\mu = \sup_{\substack{\mathbf{s}_1, \mathbf{s}_2 \in \mathcal{B}_{2\pi} \\ \mathbf{s}_1 \neq \mathbf{s}_2}} \frac{\mathrm{dist}(\mathbf{s}_1, \mathbf{s}_2)}{\lVert \mathbf{s}_2 - \mathbf{s}_1 \rVert} = 1 .
\end{align}
\end{lem}
\begin{pf}
Recall from Lemma \ref{lem:exp_coords_2d} that the Euclidean and rotational distances between any pair of exponential coordinates can be written as \eqref{eq:euclidean_distance_exp_coords_2d} and \eqref{eq:rotational_distance_exp_coords_2d} where $t \in [0,1]$.
From these equations, we can write the argument of the distance ratio constant as
\begin{align} \label{eq:exp_coords_sensitivity_fraction_2d}
\begin{aligned}
&\frac{\mathrm{dist}(\mathbf{s}_1,\mathbf{s}_2)}{\lVert \mathbf{s}_2 - \mathbf{s}_1 \rVert} \\
&\hspace{6mm}= \frac{2 \cos^{-1} \left( \lvert t \cos (\frac{\theta_1}{2} + \frac{\theta_2}{2}) + (1 - t) \cos (\frac{\theta_1}{2} - \frac{\theta_2}{2}) \rvert \right)}{\sqrt{(\theta_1 - \theta_2)^2 + 4 \theta_1 \theta_2 t}} .
\end{aligned}
\end{align}
Using convexity, we will first show that the equation above is less than or equal to one, for the case in which $\theta_1, \theta_2 \in [0,2\pi]$ and $\theta_1 + \theta_2 \in [0, 2\pi]$.

Consider the function $\cos (\sqrt{x})$.
We note that this function is convex on the interval $x \in [0,\pi^2]$, which can be verified by determining that the function's second derivative is non-negative on this interval.

So for $t \in [0,1]$, using the definition of convexity, we have
\begin{align} \label{eq:cos_sqrt_x_convex}
t \cos(\sqrt{x_1}) + (1 - t) \cos(\sqrt{x_2}) \geq \cos(\sqrt{t x_1 + (1 - t) x_2}) .
\end{align}


Considering $\theta_1$ and $\theta_2$ such that $\theta_1, \theta_2 \in [0,2\pi]$ and $\theta_1 + \theta_2 \in [0,2\pi]$, we set $x_1 = (\tfrac{\theta_1}{2} + \tfrac{\theta_2}{2})^2$ and $x_2 = (\tfrac{\theta_1}{2} - \tfrac{\theta_2}{2})^2$.
Note that $(\tfrac{\theta_1}{2} + \tfrac{\theta_2}{2})^2 \in [0,\pi^2]$ and $(\tfrac{\theta_1}{2} - \tfrac{\theta_2}{2})^2 \in [0,\pi^2]$.
Substituting these values into \eqref{eq:cos_sqrt_x_convex} and manipulating the equation algebraically yields
\begin{align}
\begin{aligned}
& t \cos(\tfrac{\theta_1}{2} + \tfrac{\theta_2}{2}) + (1-t) \cos(\tfrac{\theta_1}{2} - \tfrac{\theta_2}{2}) \\
& \hspace{26mm} \geq \cos ( \tfrac{1}{2} \sqrt{(\theta_1-\theta_2)^2 + 4 \theta_1 \theta_2 t} )
\label{eq:t_cos_x_minus_y}
\end{aligned}
\end{align}
where we have used the fact that $\cos(\sqrt{x^2}) = \cos( \lvert x \rvert) = \cos(x)$.

Next, we will apply the arccosine function to both sides of \eqref{eq:t_cos_x_minus_y}.
Note that the argument of the cosine function on the RHS of \eqref{eq:t_cos_x_minus_y} is on the interval $[0,\pi]$ and therefore applying the arccosine function to the RHS will return the original argument of the cosine function. Also, note that the LHS is on the interval $[-1,1]$ because it is a convex combination of cosine functions. Since arccosine is a decreasing function, we can apply it to both sides of \eqref{eq:t_cos_x_minus_y} and flip the inequality:
\begin{align}
\begin{aligned}
& \cos^{-1} \left( t \cos(\tfrac{\theta_1}{2} + \tfrac{\theta_2}{2}) + (1-t) \cos(\tfrac{\theta_1}{2} - \tfrac{\theta_2}{2}) \right) \\
& \hspace{33mm} \leq \frac{1}{2} \sqrt{(\theta_1-\theta_2)^2 + 4 \theta_1 \theta_2 t} .
\end{aligned}
\end{align}
Furthermore, since arccosine is a decreasing function, we have $\cos^{-1}(\lvert x \rvert) \leq \cos^{-1}(x)$ and we can write the equation above as
\begin{align}
\begin{aligned}
& 2 \cos^{-1} \left( \lvert t \cos \left( \tfrac{\theta_1}{2} + \tfrac{\theta_2}{2} \right) + (1-t) \cos \left( \tfrac{\theta_1}{2} - \tfrac{\theta_2}{2} \right) \rvert \right) \\
& \hspace{45mm} \leq \sqrt{(\theta_1-\theta_2)^2 + 4 \theta_1 \theta_2 t} .
\end{aligned}
\end{align}
Dividing by the RHS yields
\begin{align}
\begin{aligned}
& \frac{2 \cos^{-1} \left( \lvert t \cos (\frac{\theta_1}{2} + \frac{\theta_2}{2}) + (1 - t) \cos (\frac{\theta_1}{2} - \frac{\theta_2}{2}) \rvert \right)}{\sqrt{(\theta_1 - \theta_2)^2 + 4 \theta_1 \theta_2 t}} \leq 1 .
\end{aligned}
\end{align}
We can see that the expression on the LHS is the same as the RHS of \eqref{eq:exp_coords_sensitivity_fraction_2d}.
So, we have shown that the distance ratio constant is less than or equal to one for cases in which $\theta_1, \theta_2 \in [0,2\pi]$ and $\theta_1 + \theta_2 \in [0,2\pi]$.
We can complete the Lemma by showing that the distance ratio constant is less than or equal to one in cases for which $\theta_1, \theta_2 \in [0,2\pi]$ and $\theta_1 + \theta_2 \in [2\pi,4\pi]$.

Note that any points such that $\theta_1, \theta_2 \in [0,2\pi]$ and $\theta_1 + \theta_2 \in [2\pi,4\pi]$ can be written as
\begin{align}
\begin{aligned}
\theta_1 = 2\pi - \theta_1' \\
\theta_2 = 2\pi - \theta_2'
\end{aligned}
\end{align}
where $\theta_1'+\theta_2' \in [0,2\pi]$.
Plugging $(\theta_1,\theta_2,t)$ and $(\theta_1',\theta_2',t)$ into \eqref{eq:rotational_distance_exp_coords_2d} shows that their rotational distances are the same.
However, the Euclidean distance of $(\theta_1,\theta_2,t)$ is larger than that of $(\theta_1',\theta_2',t)$, which can be seen by referencing \eqref{eq:euclidean_distance_exp_coords_2d}.
Using \eqref{eq:euclidean_distance_exp_coords_2d}, we can show that the square of the Euclidean distance of $(\theta_1,\theta_2,t)$ is larger than that of $(\theta_1',\theta_2',t)$, which implies the Euclidean distance of $(\theta_1,\theta_2,t)$ is larger than that of $(\theta_1',\theta_2',t)$.
Taking the difference in the square of the Euclidean distances yields
\begin{align}
\begin{aligned}
& ( ( \theta_1 - \theta_2)^2 + 4 \theta_1 \theta_2 t ) - ( (\theta_1' - \theta_2')^2 + 4 \theta_1' \theta_2' t ) \\
& \hspace{39mm}= 8 \pi t (2 \pi - (\theta_1' + \theta_2')) .
\end{aligned}
\end{align}
Since $t$ is non-negative, and $\theta_1'+\theta_2' \in [0,2\pi]$, then the RHS of the expression above is non-negative, which implies the Euclidean distance of $(\theta_1,\theta_2,t)$ is larger than that of $(\theta_1',\theta_2',t)$. Since the Euclidean distance of $(\theta_1,\theta_2,t)$ is larger than that of $(\theta_1',\theta_2',t)$, but the rotational distance is the same for both, it follows that \eqref{eq:exp_coords_sensitivity_fraction_2d} for $(\theta_1,\theta_2,t)$ is less than or equal to that for $(\theta_1',\theta_2',t)$, which we have showed is bounded by one.

The final step is to show that \eqref{eq:exp_coords_sensitivity_fraction_2d} achieves the upper bound of one. There are a few cases in which it is achieved, one of which being when $\theta_1 \in (0, \pi]$ and $\theta_2 = 0$, in which case both $\mathrm{dist}(\mathbf{s}_1, \mathbf{s}_2)$ and $\lVert \mathbf{s}_2 - \mathbf{s}_1 \rVert$ equal $\theta_1$.
\hspace*{\fill}$\square$
\end{pf}

Next, we can use Lemma \ref{lem:drc_exp_coords_0_2pi} to derive the distance ratio constant for unconstrained exponential coordinates.
Note that the set of unconstrained exponential coordinates is all of $\mathbb{R}^3$.


\begin{thm} \label{thm:drc_uncon_exp_coords}
The distance ratio constant, $\mu$, of unconstrained exponential coordinates is:
\begin{align}
\mu = \sup_{\substack{\mathbf{s}_1, \mathbf{s}_2 \in \mathbb{R}^3 \\ \mathbf{s}_1 \neq \mathbf{s}_2}} \frac{\mathrm{dist}(\mathbf{s}_1, \mathbf{s}_2)}{\lVert \mathbf{s}_2 - \mathbf{s}_1 \rVert} = 1 .
\end{align}
\end{thm}
\begin{pf}
Recall from Lemma \ref{lem:exp_coords_2d} that the Euclidean and rotational distances between any pair of exponential coordinates can be written as
\begin{align}
\lVert \mathbf{s}_2 - \mathbf{s}_1 \rVert &{=} \sqrt{(\theta_1 - \theta_2)^2 + 4 \theta_1 \theta_2 t} \label{eq:euclidean_distance_exp_coords_2d_reproduced} \\
\mathrm{dist}(\mathbf{s}_1, \mathbf{s}_2) &{=} 2 \cos^{-1} ( | t \cos (\tfrac{\theta_1}{2}{+}\tfrac{\theta_2}{2}) + (1{-}t) \cos (\tfrac{\theta_1}{2}{-}\tfrac{\theta_2}{2}) | ) \label{eq:rotational_distance_exp_coords_2d_reproduced}
\end{align}
respectively where $t \in [0,1]$.
We will start by considering the two following cases: $\lvert \theta_1 - \theta_2 \rvert \leq \pi$ and $\lvert \theta_1 - \theta_2 \rvert > \pi$.

For the case in which $\lvert \theta_1 - \theta_2 \rvert > \pi$, upon inspection of \eqref{eq:euclidean_distance_exp_coords_2d_reproduced}, we can see that the ``$(\theta_1 - \theta_2)^2$'' term will be larger than $\pi^2$, and since $\theta_1$, $\theta_2$, and $t$ are non-negative, the $4 \theta_1 \theta_2 t$ term will be non-negative.
Therefore, the Euclidean distance will always be larger than $\pi$ which is the maximum possible rotational distance, implying that $\mathrm{dist}(\mathbf{s}_1,\mathbf{s}_2)/\lVert \mathbf{s}_2-\mathbf{s}_1 \rVert < 1$ for all points such that $\lvert \theta_1 - \theta_2 \rvert > \pi$.

For the case in which $\lvert \theta_1-\theta_2 \rvert \leq \pi$, we can shift $\theta_1$ and $\theta_2$ by a multiple of $2 \pi$ as follows (see Figure \ref{fig:theta_shift}):
\begin{align}
\begin{aligned}
\theta_1' &= \theta_1 - 2 n \pi \\
\theta_2' &= \theta_2 - 2 n \pi
\end{aligned}
\label{eq:theta_shift}
\end{align}
where $\theta_1', \theta_2' \in [-\pi,2\pi]$ and $n \in \{0,1,2,...\}$, and the larger of $\theta_1'$ and $\theta_2'$ is non-negative.
\begin{figure}[ht]
\centering
\includegraphics[width=.45\textwidth]{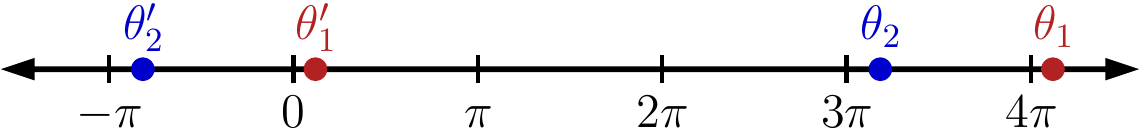}
\caption{Any angles $\theta_1$ and $\theta_2$ such that $\lvert \theta_1 - \theta_2 \rvert \leq \pi$ can be shifted by a multiple of $2\pi$ such that the resulting angles lie in the interval $[-\pi,2\pi]$, and the larger of the two is non-negative (see \eqref{eq:theta_shift}).}
\label{fig:theta_shift}
\end{figure}
Note that the numbering of $\mathbf{s}_1$ and $\mathbf{s}_2$, and $\theta_1$ and $\theta_2$, is arbitrary, so without loss of generality, we can always assume that $\theta_1' > \theta_2'$, which results in two scenarios: $\theta_1' \geq 0$ and $\theta_2' < 0$, or $\theta_1' \geq 0$ and $\theta_2' \geq 0$.
For both scenarios, we will show that $\mathrm{dist}(\mathbf{s}_1,\mathbf{s}_2)/\lVert \mathbf{s}_2-\mathbf{s}_1 \rVert$ is smaller for $(\theta_1,\theta_2,t)$ than $(\theta_1',\theta_2',t)$ which is bounded by one.

\textit{Case 1:} $\theta_1' \geq 0$ and $\theta_2' \geq 0$.
Both $(\theta_1,\theta_2,t)$ and $(\theta_1',\theta_2',t)$ correspond to the same 3D rotation, so they have the same rotational distance.
However, the Euclidean distance will be larger for $(\theta_1,\theta_2,t)$, which is apparent by inspecting \eqref{eq:euclidean_distance_exp_coords_2d_reproduced}, where we can see that the ``$(\theta_1 - \theta_2)^2$'' term will be the same for both sets of points, but the ``$4 \theta_1 \theta_2 t$'' term will be larger for $(\theta_1,\theta_2,t)$ than for $(\theta_1',\theta_2',t)$ since $0 \leq \theta_1' \leq \theta_1$ and $0 \leq \theta_2' \leq \theta_2$.
Since $(\theta_1,\theta_2,t)$ has the same rotational distance but a larger Euclidean distance than $(\theta_1',\theta_2',t)$, its value of $\mathrm{dist}(\mathbf{s}_1,\mathbf{s}_2)/\lVert \mathbf{s}_2-\mathbf{s}_1 \rVert$ is smaller than that of $(\theta_1',\theta_2',t)$ which by Lemma \ref{lem:drc_exp_coords_0_2pi} we know is less than or equal to one.

\textit{Case 2:} $\theta_1' \geq 0$ and $\theta_2' < 0$.
Both $(\theta_1,\theta_2,t)$ and $(\theta_1',\theta_2',t)$ correspond to the same 3D rotation, so they have the same rotational distance.
However, the Euclidean distance will be larger for $(\theta_1,\theta_2,t)$, which is apparent by inspecting \eqref{eq:euclidean_distance_exp_coords_2d_reproduced}, where we can see that the ``$(\theta_1 - \theta_2)^2$'' term will be the same for both sets of points, but the ``$4 \theta_1 \theta_2 t$'' term for $(\theta_1',\theta_2',t)$ will be non-positive and therefore smaller than that of $(\theta_1,\theta_2,t)$ which is non-negative.
Unlike Case 1, we cannot apply Lemma \ref{lem:drc_exp_coords_0_2pi} to $(\theta_1',\theta_2',t)$ because $\theta_2' < 0$.
However, we can note that since the axis-angle $(\theta,\mathbf{e})$ is equivalent to $(-\theta,-\mathbf{e})$, the pair $(\theta_1',\theta_2',t)$ is identical to the pair $(\theta_1',-\theta_2',1{-}t)$, where $-\theta_2' \in (0,\pi]$ and $1{-}t \in [0,1]$.
So since $(\theta_1,\theta_2,t)$ has the same rotational distance but a larger Euclidean distance than $(\theta_1',\theta_2',t)$ and $(\theta_1',-\theta_2',1{-}t)$, it has a smaller value of $\mathrm{dist}(\mathbf{s}_1,\mathbf{s}_2)/\lVert \mathbf{s}_2-\mathbf{s}_1 \rVert$ than $(\theta_1',\theta_2',t)$ and $(\theta_1',-\theta_2',1{-}t)$ which by Lemma \ref{lem:drc_exp_coords_0_2pi} we know is less than or equal to one.
\hspace*{\fill}$\square$
\end{pf}

Finally, we show that the distance ratio constant of exponential coordinates with angles $\theta \in [0,\pi]$ also equals one.
\begin{thm} \label{thm:drc_exp_coords}
The distance ratio constant, $\mu$, of exponential coordinates with angles on the interval $[0,\pi]$ is:
\begin{align}
\mu = \sup_{\substack{\mathbf{s}_1, \mathbf{s}_2 \in \mathcal{B}_{\pi} \\ \mathbf{s}_1 \neq \mathbf{s}_2}} \frac{\mathrm{dist}(\mathbf{s}_1, \mathbf{s}_2)}{\lVert \mathbf{s}_2 - \mathbf{s}_1 \rVert} = 1 .
\end{align}
\end{thm}
\begin{pf}
Lemma \ref{lem:drc_exp_coords_0_2pi} states that the distance ratio constant of exponential coordinates with angles $\theta \in [0,2\pi]$ is one. Exponential coordinates with angles $\theta \in [0,\pi]$ are a subset of those exponential coordinates, and therefore the distance ratio constant of exponential coordinates with angles $\theta \in [0,\pi]$ is upper bounded by one. Furthermore, this upper bound is achieved in a few ways, one of which being when $\theta_1 \in (0, \pi]$ and $\theta_2 = 0$, in which case both $\mathrm{dist}(\mathbf{s}_1, \mathbf{s}_2)$ and $\lVert \mathbf{s}_2 - \mathbf{s}_1 \rVert$ equal $\theta_1$.
\hspace*{\fill}$\square$
\end{pf}


\section{Quaternions} \label{sec:quaternions}

In this section, we will derive the distance ratio constant from Definition \ref{defn:distance_ratio_constant} for quaternions and unconstrained quaternions.

\subsection{Rotation parameterization} \label{sec:quaternions_rotation_parameterization}

Quaternions are a common representation of rotation which can be derived from an axis-angle representation. Considering an angle, $\theta \in \mathbb{R}$, and unit-length axis, $\mathbf{e} \in \mathbb{R}^3$, a quaternion can be written as \cite[App.~A]{Lynch}
\begin{equation}
\mathbf{q} = \begin{bmatrix} \cos \tfrac{\theta}{2} \\ \mathbf{e} \sin \tfrac{\theta}{2} \end{bmatrix} \in \mathbb{R}^4 .
\label{eq:quat_from_axis_angle}
\end{equation}

The set of all possible quaternions corresponds to all 4D vectors with unit norm, which is equivalent to all points on the 3-sphere, $\mathcal{S}^3 \subset \mathbb{R}^4$:
\begin{align} \label{eq:S3}
\mathcal{S}^3 = \{ \mathbf{x} \in \mathbb{R}^4 ~~ | ~~ \lVert \mathbf{x} \rVert = 1 \} .
\end{align}

Of particular relevance for the work here, quaternions are a double-covering of all possible 3D rotations, i.e., there are two quaternions that correspond to every 3D rotation.
More specifically, every pair of antipodal quaternions represent the same rotation, which can be seen mathematically in \eqref{eq:quat_from_axis_angle} by taking quaternion $\mathbf{q}$ with angle $\theta$, and replacing $\theta$ with the equivalent angle $\theta + 2 \pi$.
Note that while we could reduce the set $\mathcal{S}_3$ so that it is a 1-to-1 mapping, it would not change the results in this section.

Another version of quaternions we will consider are \textit{unconstrained quaternions}, which are constructed by eliminating the unit-length constraint of quaternions, thereby allowing any vector in $\mathbb{R}^4$ to be associated with a 3D rotation after normalization into a unit quaternion.
Unconstrained quaternions are advantageous as they allow a neural network to output any 4D vector as a valid parameterization of rotation.
We will consider unconstrained quaternions in Section \ref{sec:unconstrained_quaternions}.

\subsection{Rotational distance}

We can derive the rotational distance between two quaternions using the equation for the rotational distance of exponential coordinates in Proposition \ref{prp:dist_exp_coords}, which is written as a function of axis $\mathbf{e}$ and angle $\theta$.
\begin{prop}
The rotational distance between quaternions $\mathbf{q}_1, \mathbf{q}_2 \in \mathcal{S}^3$ is given by:
\begin{equation}
\mathrm{dist}(\mathbf{q}_1, \mathbf{q}_2) = 2 \cos^{-1} ( \lvert \mathbf{q}_1 \cdot \mathbf{q}_2 \rvert ) .
\label{eq:dist_quat}
\end{equation}
\end{prop}
\begin{pf}
Using \eqref{eq:quat_from_axis_angle}, we can write the quaternion vectors as
\begin{align}
\mathbf{q}_1 = \begin{bmatrix} \cos \tfrac{\theta_1}{2} \\ \mathbf{e}_1 \sin \tfrac{\theta_1}{2} \end{bmatrix}, ~~~
\mathbf{q}_2 = \begin{bmatrix} \cos \tfrac{\theta_2}{2} \\ \mathbf{e}_2 \sin \tfrac{\theta_2}{2} \end{bmatrix} .
\end{align}
Taking the dot product of $\mathbf{q}_1$ and $\mathbf{q}_2$ yields
\begin{align}
\begin{aligned}
\mathbf{q}_1 \cdot \mathbf{q}_2
&= \cos \tfrac{\theta_1}{2} \cos \tfrac{\theta_2}{2} + \mathbf{e}_1 \cdot \mathbf{e}_2 \sin \tfrac{\theta_1}{2} \sin \tfrac{\theta_2}{2} .
\end{aligned}
\label{eq:q1_dot_q2}
\end{align}
The RHS of \eqref{eq:q1_dot_q2} is exactly the expression in the absolute value in \eqref{eq:dist_axang}, which is the equation for the rotational distance between axis-angle pairs ($\theta_1$,$\mathbf{e}_1$) and ($\theta_2$,$\mathbf{e}_2$). Since $\mathbf{q}_1$ and $\mathbf{q}_2$ correspond to the same 3D rotations as ($\theta_1$,$\mathbf{e}_1$) and ($\theta_2$,$\mathbf{e}_2$), respectively, plugging \eqref{eq:q1_dot_q2} into \eqref{eq:dist_axang} yields \eqref{eq:dist_quat}.
\hspace*{\fill}$\square$
\end{pf}

We can also express the rotational distance between two quaternions as a function of the angle between their 4D quaternion vectors. This result will be used in Section \ref{sec:quaternion_2d_interpretation}.
\begin{prop}
Consider two quaternions $\mathbf{q}_1, \mathbf{q}_2 \in \mathcal{S}^3$, and let $\phi$ denote the angle between the 4D quaternion vectors. The rotational distance between the quaternions can be expressed in terms of $\phi$ as:
\begin{equation}
\mathrm{dist}(\mathbf{q}_1, \mathbf{q}_2)
=
\begin{cases}
2 \phi, & \phi \in [0, \pi/2] \\
2 \pi - 2 \phi, & \phi \in [\pi/2, \pi] .
\end{cases}
\label{eq:dist_quat_phi}
\end{equation}
\label{prp:dist_quat_phi}
\end{prop}
\begin{pf}
Applying the formula which relates the dot product and angle between two vectors, and noting that $\mathbf{q}_1$ and $\mathbf{q}_2$ are unit length, we have  
\begin{align}
\begin{aligned}
\cos \phi &= \frac{\mathbf{q}_1 \cdot \mathbf{q}_2}{\lVert \mathbf{q}_1 \rVert \lVert \mathbf{q}_2 \rVert} = \mathbf{q}_1 \cdot \mathbf{q}_2 .
\end{aligned}
\label{eq:cos_phi}
\end{align}
Plugging \eqref{eq:cos_phi} into \eqref{eq:dist_quat} yields
\begin{equation}
\mathrm{dist}(\mathbf{q}_1, \mathbf{q}_2) = 2 \cos^{-1} ( \lvert \cos \phi \rvert ) .
\label{eq:dist_quat_phi_arccos_cos}
\end{equation}
Now note the following property
\begin{equation}
\cos^{-1} ( \lvert \cos \phi \rvert )
=
\begin{cases}
\phi, & \phi \in [0, \pi/2] \\
\pi - \phi, & \phi \in [\pi/2, \pi] .
\end{cases}
\label{eq:arccos_abs_cos_property}
\end{equation}
Plugging \eqref{eq:arccos_abs_cos_property} into \eqref{eq:dist_quat_phi_arccos_cos} yields \eqref{eq:dist_quat_phi}.
\hspace*{\fill}$\square$
\end{pf}


\subsection{Planar analysis} \label{sec:quaternion_2d_interpretation}

\begin{figure}[ht]
\centering
\includegraphics[width=.40\textwidth]{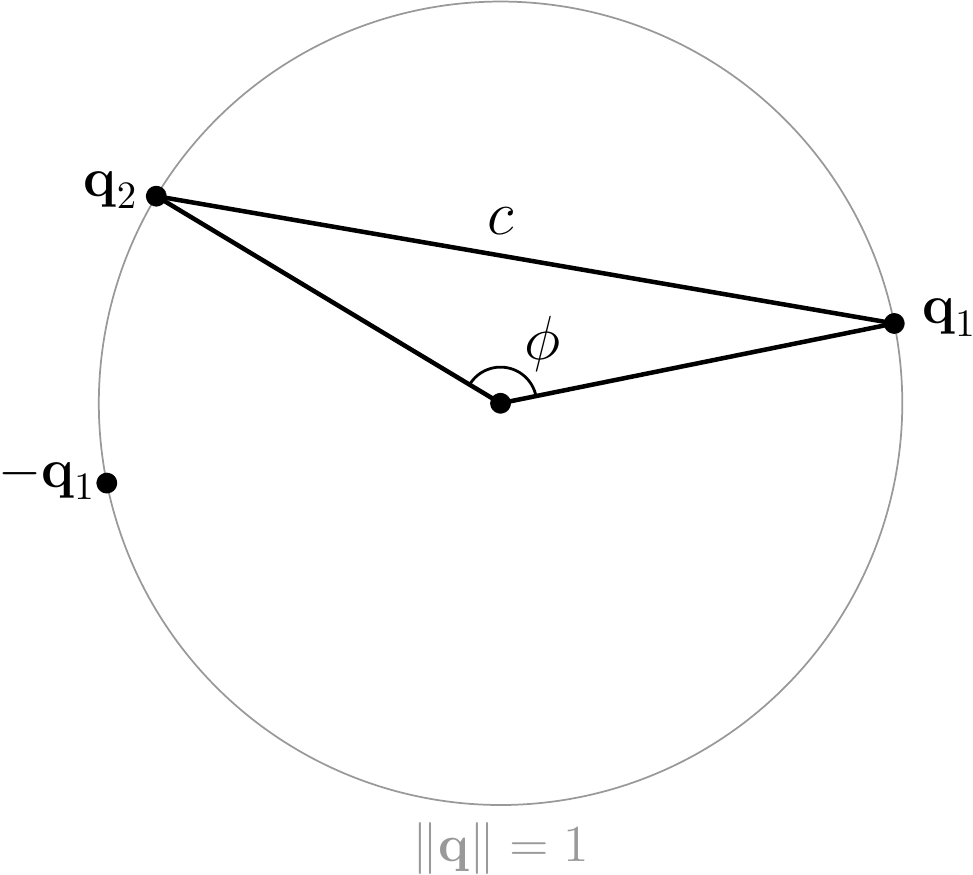}
\caption{
Quaternions $\mathbf{q}_1$ and $\mathbf{q}_2$ shown in the plane they define.
In the plane, two quaternions can be analyzed in terms of the angle $\phi$ between the vectors, and the chord length $c$, which is equivalent to their Euclidean distance $\lVert \mathbf{q}_2 - \mathbf{q}_1 \rVert$.
Note that when the angle $\phi$ is larger than $\pi/2$, $\mathbf{q}_2$ will be closer to $-\mathbf{q}_1$ than $\mathbf{q}_1$.
}
\label{fig:quaternion_circle}
\end{figure}

To determine the distance ratio constant of quaternions, we need to optimize over all possible pairs of quaternions, where each pair consists of the eight degrees of freedom in $\mathbf{q}_1$ and $\mathbf{q}_2$, and the two constraints $\lVert \mathbf{q}_1 \rVert = 1$ and $\lVert \mathbf{q}_2 \rVert = 1$.
The number of dimensions and constraints in this analysis is challenging, but we can analyze this problem in the plane defined by $\mathbf{q}_1$ and $\mathbf{q}_2$, which reduces the number of variables and simplifies the analysis.

\begin{lem} \label{lem:quaternion_2d}
Consider two quaternions $\mathbf{q}_1, \mathbf{q}_2 \in \mathcal{S}^3$, and let $c$ denote their Euclidean distance. The ratio of the rotational distance and the Euclidean distance between the two quaternions is given by:
\begin{align}
\frac{\mathrm{dist}(\mathbf{q}_1, \mathbf{q}_2)}{\lVert \mathbf{q}_2 - \mathbf{q}_1 \rVert}
= \begin{cases}
4 \sin^{-1}(\tfrac{c}{2})/c, &c \in (0, \sqrt{2}] \\
(2 \pi - 4 \sin^{-1}(\tfrac{c}{2}))/c, &c \in [\sqrt{2}, 2] .
\end{cases}
\label{eq:sens_frac_c}
\end{align}
\end{lem}
\begin{pf}
We start by noting that any two quaternions define a plane, and hence we can analyze their relationship in that plane.
A diagram is shown in Fig. \ref{fig:quaternion_circle}. Since $\mathbf{q}_1$ and $\mathbf{q}_2$ lie on the unit circle, the distance between them can be interpreted as a chord length.
Denoting this chord length as $c$, we have
\begin{equation}
c = \lVert \mathbf{q}_2 - \mathbf{q}_1 \rVert .
\label{eq:quat_euclidean_distance_c}
\end{equation}
Let $\phi$ denote the angle between the $\mathbf{q}_1$ vector and the $\mathbf{q}_2$ vector. We can relate the chord length $c$, angle $\phi$, and radius $\rho$ using the trigonometric formula $\phi = 2 \sin^{-1}(\tfrac{c}{2 \rho})$. Since quaternions are unit-length, in this case $\rho = 1$, and we can write the relationship between $c$ and $\phi$ as
\begin{align}
\phi = 2 \sin^{-1}(\tfrac{c}{2}) .
\label{eq:phi_function_of_c}
\end{align}
Plugging \eqref{eq:phi_function_of_c} into \eqref{eq:dist_quat_phi}, and converting the intervals accordingly, yields
\begin{equation}
\mathrm{dist}(\mathbf{q}_1, \mathbf{q}_2)
=
\begin{cases}
4 \sin^{-1}(\tfrac{c}{2}), & c \in (0, \sqrt{2}] \\
2 \pi - 4 \sin^{-1}(\tfrac{c}{2}), & c \in [\sqrt{2}, 2] .
\end{cases}
\label{eq:dist_quat_c}
\end{equation}
Plugging \eqref{eq:dist_quat_c} and \eqref{eq:quat_euclidean_distance_c} into the LHS of \eqref{eq:sens_frac_c} yields the RHS of \eqref{eq:sens_frac_c}.
Note that in \eqref{eq:sens_frac_c} we do not consider $c=0$ as it corresponds to $\mathbf{q}_1 = \mathbf{q}_2$, which is not considered in the distance ratio constant from Definition \ref{defn:distance_ratio_constant}.
\hspace*{\fill}$\square$
\end{pf}

\subsection{Distance ratio constant}

Next, we will derive the distance ratio constant.
To facilitate the analysis, we will use the planar interpretation from Section \ref{sec:quaternion_2d_interpretation}.
\begin{thm} \label{thm:drc_quat}
The distance ratio constant, $\mu$, of quaternions $\mathbf{q}_1$ and $\mathbf{q}_2$ is:
\begin{align} \label{eq:rotational_sensitivity_constant_quaternions}
\mu = \sup_{\substack{\mathbf{q}_1, \mathbf{q}_2 \in \mathcal{S}^3 \\ \mathbf{q}_1 \neq \mathbf{q}_2}} \frac{\mathrm{dist}(\mathbf{q}_1, \mathbf{q}_2)}{\lVert \mathbf{q}_2 - \mathbf{q}_1 \rVert} = \frac{\pi}{\sqrt{2}} .
\end{align}
\end{thm}
\begin{pf}
In Lemma \ref{lem:quaternion_2d} we showed that we can write the sensitivity of any two quaternions in terms of the single variable $c$, which represents their Euclidean distance and chord length in the plane defined by the two quaternion vectors.
Taking the derivative of \eqref{eq:sens_frac_c} with respect to $c$ yields
\begin{align}
\frac{d}{dc} \frac{\mathrm{dist}(\mathbf{q}_1, \mathbf{q}_2)}{\lVert \mathbf{q}_2 - \mathbf{q}_1 \rVert} {=}
\begin{cases}
\frac{4}{c^2} \left( \frac{c}{\sqrt{4-c^2}} - \sin^{-1}(\tfrac{c}{2}) \right), &\hspace{-3.1mm}c \in (0, \sqrt{2}] \label{eq:quat_sens_deriv} \\
\frac{-4}{c^2} \left( \frac{c}{\sqrt{4-c^2}} + \cos^{-1}(\tfrac{c}{2}) \right), &\hspace{-3.1mm}c \in [\sqrt{2}, 2] .
\end{cases}
\end{align}

To begin, consider the first case of \eqref{eq:quat_sens_deriv}, where $c \in (0,\sqrt{2}]$. The second term can be rearranged as follows:
\begin{align}
\begin{aligned}
\frac{c}{\sqrt{4-c^2}} - \sin^{-1}(\tfrac{c}{2})
&= (\tfrac{c}{2})/\sqrt{(1-(\tfrac{c}{2}))^2} - \sin^{-1}(\tfrac{c}{2}) \\
&= \tan (\sin^{-1}(\tfrac{c}{2})) - \sin^{-1}(\tfrac{c}{2})
\end{aligned}
\label{eq:sens_deriv_c_case_1}
\end{align}
where in the last step we used the identity $\tan (\sin^{-1} x) = x / \sqrt{1 - x^2}$.
Note that $c/2$ is positive, so $\sin^{-1} (\tfrac{c}{2})$ is on the interval $(0,\pi/2]$.
Because $\tan x > x$ for $x \in (0, \pi/2]$, the bottom equation in \eqref{eq:sens_deriv_c_case_1} is positive, which since $c > 0$, implies that \eqref{eq:quat_sens_deriv} is positive in the case of $c \in (0, \sqrt{2}]$.

Next, consider the second case of \eqref{eq:quat_sens_deriv}, where $c \in [\sqrt{2},2]$.
Restricting the interval to $[\sqrt{2},2)$, the second term consists of the sum of two non-negative terms, which is non-negative.
Since $-4/c^2$ is negative, \eqref{eq:quat_sens_deriv} is non-positive in the case of $c \in [\sqrt{2},2)$.


In summary, the derivative of \eqref{eq:sens_frac_c} with respect to $c$ is positive for $c \in (0, \sqrt{2}]$ and non-positive for $c \in [\sqrt{2}, 2)$, so \eqref{eq:sens_frac_c} is maximized at $c = \sqrt{2}$ on the interval $c \in (0,2)$.
Evaluating \eqref{eq:sens_frac_c} with $c = \sqrt{2}$ yields $\pi/\sqrt{2}$.
As \eqref{eq:sens_frac_c} equals zero when $c=2$, it has a maximum value of $\pi/\sqrt{2}$ on the interval $(0,2]$.
\hspace*{\fill}$\square$
\end{pf}

\subsection{Unconstrained quaternions} \label{sec:unconstrained_quaternions}

We now consider unconstrained quaternions, which as we mentioned in Section \ref{sec:quaternions_rotation_parameterization}, are constructed by eliminating the unit-length constraint of quaternions, thereby allowing any vector in $\mathbb{R}^4$ to be associated with a 3D rotation after normalization into a unit quaternion.
We now derive the distance ratio constant of unconstrained quaternions.

\begin{thm} \label{thm:drc_uncon_quat}
The distance ratio constant, $\mu$, of unconstrained quaternions $\mathbf{q}_1$ and $\mathbf{q}_2$ is:
\begin{align}
\mu = \sup_{\substack{\mathbf{q}_1, \mathbf{q}_2 \in \mathbb{R}^4 \\ \mathbf{q}_1 \neq \mathbf{q}_2}} \frac{\mathrm{dist}(\mathbf{q}_1, \mathbf{q}_2)}{\lVert \mathbf{q}_2 - \mathbf{q}_1 \rVert} = \infty .
\end{align}
\end{thm}
\begin{pf}
Consider any two unconstrained quaternions $\mathbf{q}_1, \mathbf{q}_2 \in \mathbb{R}^4$ which do not correspond to the same rotation. Let $\alpha > 0$ denote their rotational distance. Now consider the scalar, $\epsilon$, and scalar multiples of $\mathbf{q}_1$ and $\mathbf{q}_2$: $\epsilon \mathbf{q}_1$ and $\epsilon \mathbf{q}_2$. Since scaling an unconstrained quaternion does not change its associated rotation, we have
\begin{align}
\alpha = \mathrm{dist}(\mathbf{q}_1, \mathbf{q}_2)
= \mathrm{dist}(\epsilon \mathbf{q}_1, \epsilon \mathbf{q}_2) > 0 .
\end{align}
Taking the limit as $\epsilon$ goes to zero, we have
\begin{align}
\lim_{\epsilon \rightarrow 0} \frac{\mathrm{dist}(\epsilon \mathbf{q}_1, \epsilon \mathbf{q}_2)}{\lVert \epsilon \mathbf{q}_2 - \epsilon \mathbf{q}_1 \rVert} = \frac{\alpha}{0} = \infty .
\end{align}
\\[-4mm]\hspace*{\fill}$\square$
\end{pf}

Mapping an unconstrained quaternion to a unit quaternion is performed using a unit-normalization function.
We now calculate the Euclidean Lipschitz constant of a unit-normalization function.
\begin{prop} \label{prp:unit_normalization_lipschitz}
Let $\mathbf{f}$ denote the unit-normalization function. The Euclidean Lipschitz constant of $\mathbf{f}$ is $\infty$:
\begin{equation}
\sup_{\mathbf{x}_1 \neq \mathbf{x}_2} \frac{\lVert \mathbf{f}(\mathbf{x}_2) - \mathbf{f}(\mathbf{x}_1) \rVert}{\lVert \mathbf{x}_2 - \mathbf{x}_1 \rVert} 
= \infty .
\label{eq:unit_normalization_lipschitz_constant}
\end{equation}
\end{prop}
\begin{pf}
To prove this proposition, we can show that the fraction in \eqref{eq:unit_normalization_lipschitz_constant} approaches infinity for some vectors $\mathbf{x}_1$ and $\mathbf{x}_2$.
Consider two unit vectors $\mathbf{u}_1 \neq \mathbf{u}_2$. Let $\epsilon \mathbf{u}_1$ and $\epsilon \mathbf{u}_2$ denote scalings of these vectors by $\epsilon > 0$. Since $\mathbf{f}$ is a normalization function, we have $\mathbf{f}(\epsilon \mathbf{u}_1) = \mathbf{u}_1$ and $\mathbf{f}(\epsilon \mathbf{u}_2) = \mathbf{u}_2$. Let $b$ denote the Euclidean distance between $\mathbf{f}(\mathbf{x}_1)$ and $\mathbf{f}(\mathbf{x}_2)$:
\begin{align}
b = \lVert \mathbf{f}(\epsilon \mathbf{u}_2) - \mathbf{f}(\epsilon \mathbf{u}_1) \rVert = \lVert \mathbf{u}_2 - \mathbf{u}_1 \rVert > 0 .
\end{align}
The fraction in \eqref{eq:unit_normalization_lipschitz_constant} for vectors $\epsilon \mathbf{u}_1$ and $\epsilon \mathbf{u}_2$ as $\epsilon$ approaches 0 is
\begin{equation}
\lim_{\epsilon \rightarrow 0} \frac{\lVert \mathbf{f}(\epsilon \mathbf{u}_2) - \mathbf{f}(\epsilon \mathbf{u}_1) \rVert}{\lVert \epsilon \mathbf{u}_2 - \epsilon \mathbf{u}_1 \rVert} 
= \frac{b}{0}
= \infty .
\end{equation}
\\[-4mm]\hspace*{\fill}$\square$
\end{pf}

\section{Construction of a pose estimation network with provable sensitivity bounds} \label{sec:summary_and_simulation}

In the previous two sections, we derived the distance ratio constant for several rotation parameterizations.
In this section, we will consider the task of using these results along with Theorem \ref{thm:network_wide_bound} to design a pose estimation network with provable sensitivity bounds.
We will show that unconstrained exponential coordinates are the only rotation parameterization we have considered which can accomplish this task.
We will then design and train a pose estimation network using unconstrained exponential coordinates, and compute sensitivity bounds for it.

\subsection{Summary of rotation parameterizations}

In summary, we have derived the distance ratio constant for exponential coordinates and quaternions, as well as unconstrained versions of both of these parameterizations. The results are summarized in Fig. \ref{fig:sup_summary_table}.
\begin{figure}[ht]
\setlength{\tabcolsep}{5pt}
\centering
\begin{tabular}{ l c c c }
\toprule
\textbf{parameterization} &
\textbf{symbol} ($\mathbf{p}$) &
\textbf{set} ($\mathcal{P}$) &
$\mu$ \\
\midrule
exp coords                      & $\mathbf{s}$ & $\mathcal{B}_{\pi}$ & 1 \\
exp coords, uncon.                  & $\mathbf{s}$ & $\mathbb{R}^3$ & 1 \\
quaternions                     & $\mathbf{q}$ & $\mathcal{S}^3$ & $\pi/\sqrt{2}$ \\
quaternions, uncon.                 & $\mathbf{q}$ & $\mathbb{R}^4$ & $\infty$ \\
\bottomrule
\end{tabular}
\caption{
Parameterizations, symbols, set of parameters, and distance ratio constant, $\mu$, for various rotation parameterizations that we have derived in this paper (see Theorems \ref{thm:drc_uncon_exp_coords}, \ref{thm:drc_exp_coords}, \ref{thm:drc_quat}, \& \ref{thm:drc_uncon_quat}).
The distance ratio constant, $\mu$, is defined in Definition \ref{defn:distance_ratio_constant}, and ``uncon.'' denotes ``unconstrained''.
}
\label{fig:sup_summary_table}
\end{figure}
We also verified these results to be accurate by randomly sampling $10^7$ pairs of 3D rotations and calculating the argument of the distance ratio constant for each pair.

We now consider the task of constructing a pose estimation network using each of these parameterizations, and calculating the rotational Lipschitz constant for each network.
Unfortunately, this task is unrealizable for most of these parameterizations.
Using exponential coordinates would require that the neural network only output vectors with magnitude less than or equal to $\pi$, which is not straightforward to accomplish.
Similarly, using quaternions would require the neural network to only output vectors with unit magnitude.
This unit magnitude constraint could be enforced using a unit-normalization function, but as we showed in Proposition \ref{prp:unit_normalization_lipschitz}, the unit-normalization function has an infinite Lipschitz constant.
Similarly, unconstrained quaternions have an unbounded distance ratio constant.




However, unconstrained exponential coordinates do not have any of these challenges as they associate any vector in $\mathbb{R}^3$ with a valid 3D rotation, and they have a bounded distance ratio constant.
As a result, it is possible to build a pose estimation neural network which outputs rotation in the form of an unconstrained three-dimensional vector, and then determine a bound on the network's rotational Lipschitz constant using Theorem \ref{thm:network_wide_bound}.

\subsection{A pose estimation network with provable sensitivity bounds} \label{sec:simulation}

In this section, we will design and train a neural network which outputs rotation in the form of unconstrained exponential coordinates, and then calculate a bound on its rotational Lipschitz constant.
Our network outputs six-dimensional vectors, in which the first three states represent position and the last three states represent unconstrained exponential coordinates.

\begin{figure}[ht]
\centering
\includegraphics[width=.47\textwidth]{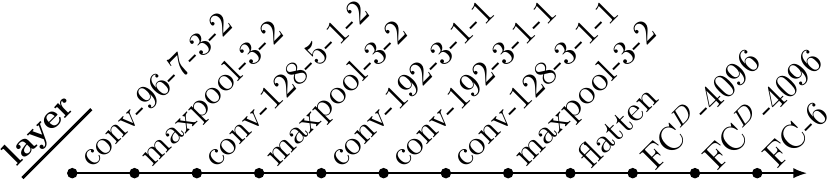}
\caption{Architecture of our neural network with provable sensitivity bounds, shown as a sequence of layers which are applied left-to-right. The layers are denoted as follows: conv-$o$-$k$-$s$-$p$ denotes a convolution layer with $o$ output channels, kernel size $k$, stride $s$ and padding $p$; maxpool-$k$-$s$ denotes a max pooling function with a kernel size of $k$ and stride $s$; and FC-$o$ denotes a fully-connected function with $o$ output channels. The superscript $D$ denotes that dropout is applied after the layer. All convolution and fully-connected layers, except for the final layer, have a ReLU activation.}
\label{fig:network_architecture}
\end{figure}

We constructed a feedforward network with rectified linear unit (ReLU) activation functions, which has a similar architecture to the original AlexNet \cite{Krizhevsky}. The network architecture is shown in Fig. \ref{fig:network_architecture}.

\begin{figure}[ht]
\centering
\includegraphics[width=.15\textwidth]{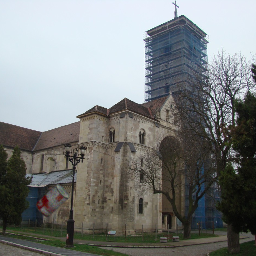}
\includegraphics[width=.15\textwidth]{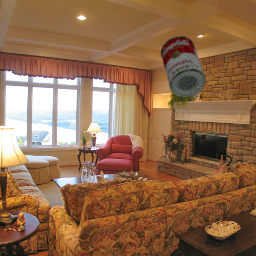}
\includegraphics[width=.15\textwidth]{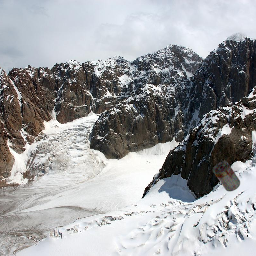}
\caption{
Sample images used to train the pose estimation neural network used in the simulation.
The object of interest is the ``soup can'' from the YCB dataset.
The images were generated synthetically by rendering the object at a random pose using Blender, and then overlaying the render onto a random background image.
}
\label{fig:training_images}
\end{figure}

We consider our object of interest to be the ``soup can'' object from the YCB data set.
We generated each image in the training and test data by generating renderings of the soup can at random poses using the computer graphics program Blender, and then superimposing the rendering on a random background image.
Our training data consisted of 200,000 renderings superimposed on random background images from the SUN397 dataset.
Our test data consisted of 10,000 renderings superimposed on background images from the 2017 COCO dataset.
Each image was of size $3{\times}256{\times}256$, and sample images are shown in Fig. \ref{fig:training_images}. 

We created the cost function for the network as follows. Let $\mathbf{z}_i \in \mathbb{R}^3$ and $\hat{\mathbf{z}}_i \in \mathbb{R}^3$ denote the true and estimated position for the $i^{th}$ data point, respectively. Let $\mathbf{s}_i \in \mathcal{B}_{\pi}$ denote the true exponential coordinates, and $\hat{\mathbf{s}}_i \in \mathbb{R}^3$ denote the estimated unconstrained coordinates for the $i^{th}$ data point. Letting $m$ denote the number of data points, we used the following cost function:
\begin{equation}
\textrm{cost} = \sum_{i=1}^m \lVert \mathbf{z}_i - \hat{\mathbf{z}}_i \rVert^2 + \mathrm{dist}(\mathbf{s}_i, \hat{\mathbf{s}}_i)
\end{equation}
where $\mathrm{dist}(\cdot,\cdot)$ is the rotational distance equation from \eqref{eq:dist_axang}.

We trained the network using the Adadelta optimization algorithm with an initial learning rate of $10^{-1}$, and reduced the learning rate to $10^{-2}$ when the error plateaued.
We trained the network for 235 epochs, which resulted in average test sample rotation and position errors of 13$\degree$ and 7cm, respectively.
Note that about 1\% of the network's exponential coordinate estimates had a norm larger than $\pi$ (i.e., were outside of the set $\mathcal{B}_{\pi}$) for both training and test data.


The network outputs a six-dimensional vector which represents position and exponential coordinates.
In order to analyze the position and rotation portions of the network independently, we split the network into two sub-networks by splitting the final fully-connected layer into two parts.
We determined upper bounds on the Euclidean Lipschitz constants of the position and rotation sub-networks to be $13 \times 10^9$ and $84 \times 10^9$, respectively.
Using the latter value along with Theorems \ref{thm:network_wide_bound} and \ref{thm:drc_uncon_exp_coords}, we determined the following upper bound on the rotational Lipschitz constant of the network:
\begin{align}
L_r
\leq 84 \times 10^9 .
\label{eq:simulation_sensitivity}
\end{align}
Note that all of our computed Lipschitz bounds are very large, but are of similar magnitude to classification networks with comparable architectures.


Finally, we can use Corollary \ref{cor:distance_bound} to determine a bound on the rotational change of the output, given a bound on input perturbations.
Using \eqref{eq:distance_bound}, we can determine that if the Euclidean distance of two inputs is less than or equal to $\epsilon = 1.1 \times 10^{-11}$, then the rotational distance between the outputs will be less than or equal to 1 radian.
We could also determine this bound for different values of $\epsilon$. 



\section{Conclusion} \label{sec:conclusion}

In this paper, we approached the task of deriving sensitivity bounds for pose estimation neural networks. 
We created a sensitivity measure that is a type of Lipschitz constant which measures the maximum rotational distance between two outputs of a network with respect to the Euclidean distance between the corresponding inputs.
We then showed how a bound on this measure can be calculated from the Euclidean Lipschitz constant of the network and the distance ratio constant of the rotation parameterization.
We derived the distance ratio constant for various rotation parameterizations, and discussed why most of these parameterizations, except for unconstrained exponential coordinates, make it difficult to compute a network-wide bound.
We then constructed a neural network which outputs pose as a concatenation of position and unconstrained exponential coordinates, and calculated a bound on the rotational Lipschitz constant of the network. 


There are several useful directions of future work we can consider.
One is to further analyze 2D projection parameterizations.
These parameterizations have shown to be very effective when used in pose estimation neural networks, and are also versatile due to the fact that these networks can be applied to images with a variety of camera intrinsics.

Finally, we again note that our sensitivity bounds were very large, but are of comparable magnitude to sensitivity bounds for classification networks.
The high sensitivity of these bounds is a direct result of the bound on the Euclidean Lipschitz constant of the network being large.
Therefore, in order to decrease our bounds and make them more practical, it is important to develop a way to calculate tighter bounds on the Euclidean Lipschitz constant of a network, or to constrain the Euclidean Lipschitz constant of the network during training.
Note that both of these topics are actively being researched and have wide-ranging applicability in the field of deep learning.


\bibliographystyle{ieeetr}
\bibliography{root.bib}

\end{document}